\documentclass[letterpaper, 10 pt, conference]{ieeeconf}  % Comment this line out if you need a4paper

\IEEEoverridecommandlockouts                              % This command is only needed if 
\usepackage{amsmath,amsfonts}
\usepackage[linesnumbered,ruled,vlined]{algorithm2e}
\usepackage{algorithmic}
\usepackage{graphicx}
\usepackage{textcomp}
\usepackage{xcolor}
\usepackage{caption} % for "\captionof" macro
\usepackage{subfigure}
\usepackage{balance}
\usepackage{mathabx}
\usepackage{blindtext}

\newtheorem{definition}{Definition}[section]

\DeclareMathOperator*{\argmaxpi}{argmax}
\DeclareMathOperator*{\argminpi}{argmin}
\title{\LARGE \bf
Generating Active Explicable Plans in
Human-Robot Teaming
}

\author{Akkamahadevi Hanni and Yu Zhang% <-this % stops a space
\thanks{The authors are with the School of Computing, Informatics, and Decision Systems Engineering at Arizona State University, Tempe, Arizona, USA.
{\tt\small \{ahanni, yzhan442\}@asu.edu}.}% <-this % stops a space
}

\begin{document}

\maketitle
\thispagestyle{empty}
\pagestyle{empty}

%%%%%%%%%%%%%%%%%%%%%%%%%%%%%%%%%%%%%%%%%%%%%%%%%%%%%%%%%%%%%%%%%%%%%%%%%%%%%%%%
\begin{abstract}
Intelligent robots are redefining a multitude of critical domains but are still far from being fully capable of assisting human peers in day-to-day tasks. 
An important requirement of collaboration is for each teammate to maintain and respect an understanding of the others' expectations of itself.
Lack of which may lead to serious issues such as loose coordination between teammates, reduced situation awareness, and ultimately teaming failures. 
Hence, it is important for robots to behave explicably by meeting the human's expectations. 
One of the challenges here is that the expectations of the human are often hidden and can change dynamically as the human interacts with the robot. 
However, existing approaches to generating explicable plans  often assume that the human's expectations are known and static. 
In this paper, we propose the idea of {\bf active explicable planning} to relax this assumption. 
We apply a Bayesian approach to model and predict dynamic human belief and expectations to make explicable planning more anticipatory. 
We hypothesize that active explicable plans can be more efficient and explicable at the same time, when compared to explicable plans generated by the existing methods. 
{\color{black} In our experimental evaluation, we verify that our approach generates more efficient explicable plans while successfully capturing the dynamic belief change of the human teammate.}
\end{abstract}

%%%%%%%%%%%%%%%%%%%%%%%%%%%%%%%%%%%%%%%%%%%%%%%%%%%%%%%%%%%%%%%%%%%%%%%%%%%%%%%%
\section{Introduction}

Advancements in robotics has opened up many opportunities for novel robotic applications. 
For example, robots can play a vital role in helping humans carry out everyday tasks such as house chores %warehouse goods management 
and accomplishing critical missions like urban search and rescue.
 In many cases, it is desirable for the robots to be fully autonomous and minimize human intervention.
As a result, a desirable capability is to have these robots maintain awareness of human presence and their expectations of the robot in the environment. Absence of this capability may lead to serious consequences such as reduced team situation awareness and loss of trust. 

The authors in~\cite{Zhang2017PlanEA} first proposed to incorporate the human's expectation in the robot's decision-making.
One of the key assumptions there is that humans generate their expectations of the robot based on a model of expectation in the human's mind,
much like how the robot generates behaviors using its domain model. 
As a result, the robot should make an effort to align its behavior with the human's expectation. 
A similar idea is adopted in other work, such as in ~\cite{Kulkarni2016ExplicableRP} and~\cite{Kulkarni2019AUF}.
% but the authors instead use plan distance measures for predicting the human's expectations.
% %both of which assuming a static human model. 
% Authors in~\cite{Kulkarni2019AUF} study a similar problem under the notion of plan legibility. 
% Along the same road, under the pretext of Plan Legibility, \cite{Kulkarni2019AUF} generate legible plans which, by characteristic, is consistent with the observations made by the human teammate. 
However, these prior approaches all assume that the model for generating human expectations remains fixed throughout the task. 
While such an assumption maybe innocent to make in scenarios where the entire plan is presented to the human at once, 
it will definitely fail to hold in human-robot teaming scenarios where the human observes the robot's actions and revise the expectations of the robot over time. 
% The focus of our work is closely related to these literatures in a way that it is laid out in the same setting but addresses one very important limitation of these earlier work: while the intent may be considered to be known and fixed within a task, the human's understanding of the robot's model can be much more dynamic. The human's understanding of a robot may come from prior domain knowledge or observing the appearance of a robot or observing a robot perform actions. 
For example, if a human saw a robot fly, the robot would be expected to fly afterwards when needed, even if the human did not know about it in the first place. 
% looks at a robot with wheels, the human may assume that it is a mobile robot which can move around but it is very unlikely that the human would assume that the robot could fly or pick up objects with a gripper. Such a bias would cause the human to expect certain behaviour from the robot which, inherently, may not be the same as the robot's intended behaviour.

\begin{table}
    \centering
    \includegraphics[scale=0.19]{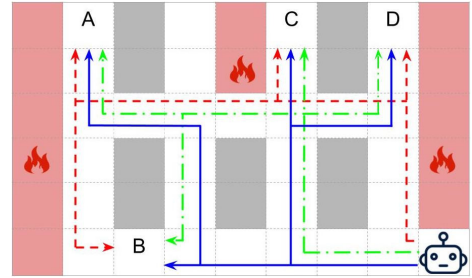} 
    \captionof{figure}{USAR domain: building layout with the paths that correspond to the optimal plan - OP (Green), explicable plan - EXP (Red) and active explicable plan - ActiveEXP (Blue).} 
    \label{fig:usar}
    \vspace{7pt}
    %\resizebox{\linewidth}{!}
    {%
    \begin{tabular}{ |c|c|c| }
     \hline
        OP (Green) & EXP (Red) & ActiveEXP (Blue)\\
     \hline
        Collect C & Collect D & Collect C \\
        Collect D & Collect C & Collect D \\
        Collect A & Collect A & Collect B \\
        Collect B & Collect B & Collect A \\ 
     \hline
    \end{tabular} }
    \captionof{table}{Steps of the three different plans in Fig. \ref{fig:usar}}
     \label{tab:plans-tab}
     \vspace{-10pt}
\end{table}
% \vspace{-10pt}

% The underlying assumption in earlier work on explicability using model based methods is that the robot has access to the human's model beforehand.
% In reality, such information is difficult to obtain without explicitly querying a human, for his/her expectations of the robot. Also, this implies that the human's expectations of the robot are known apriori and do not change during the course of plan execution. This is unlikely because the human's belief changes while observing the robot which also causes the expectations to change (which is unaccounted in existing methods). While the existing approaches focus on generating task specific plans and therefore, are designed to work on short term based plan solutions, they will undoubtedly fail in long-term interactions due to the dynamically changing expectations of the human. Therefore, maintaining a dynamic human belief in such cases becomes critical for teaming effectiveness. 

In this paper, we introduce the notion of active explicable planning\footnote{This work was first presented as a Late-Breaking Report at HRI 2021. Here, we significantly extend the work to include the derivation of the Bayesian formulation and solution method, as well as a more comprehensive result section that includes both synthetic and human subject evaluations.} to relax this assumption. 
We formulate the problem in a Bayesian framework by modeling
changes of the human's model and expectations in a Dynamic Bayesian Network.
We then derive a new explicable planning solution based on the changes  anticipated in the human's expectations throughout the plan, assuming that the human always observes the robot's actions.
% However, directly querying this network is expensive. 
% Instead, we develop an approximation  
% for an efficient solution.
{\color{black} In our experiments, we show that the plan generated by our approach is simultaneously more efficient and explicable. 
Results also confirm that humans do update their model and expectations dynamically.

%therefore, being more explicable. Interestingly, this plan inadvertently, provides clarity on hidden information in the domain.
}
%We apply a Bayesian approach to model and predict the human's dynamic belief to capture the expectations of the human at any given time. Allow us to explain with a motivating example.

The contribution of our work is three-fold: first, we introduce active explicable planning to relax a restrictive assumption of the existing work. Second, we derive a novel Bayesian formulation for active explicable planning and propose a solution.
Third, we evaluate the proposed method to demonstrate its effectiveness in human-robot teaming. 
%method for dynamic modeling of the human belief, 2) an active explicable planning framework that generates active explicable plans that are simultaneously more efficient, 3) empirical evaluation of optimal, explicable and active explicable plans. 

\subsection{Motivational Example}
Consider yourself working with a robot in an urban search and rescue scenario, which involves retrieving valuables (A, B, C and D) from a building under fire (see Fig. $1$). While the robot's task is to retrieve valuables from the building safely, your task is to supervise the robot. Assume you have information about the internal layout of the building, locations of the valuables, and  areas in the building on fire. The plan you would expect the robot to execute may be the shortest-path plan, i.e., the one shown in Red (Collect D, Collect C, Collect A and Collect B). 
This is shown also as the EXP plan in Table \ref{tab:plans-tab}. 
However, the robot could be sensitive to high temperatures and would prefer to avoid passing through \textit{long corridors} near a fire. In such cases, the optimal plan for the robot would be the path indicated in Green (Collect C, Collect D, Collect A and Collect B), shown as the OP plan. 
%would not be expected by you. 

Now consider another plan indicated in Blue (Collect C, Collect D, Collect B and Collect A), the ActiveEXP plan. When the robot first collects C, it is likely that you would consider that the robot is trying to avoid the corridor near the fire. 
You would then consider navigating near a fire as undesirable.
As such, the behavior of the robot to avoid the fire passage near C by moving around it to collect B becomes expected. %this behaviour is no longer unexpected because this path doesn't encounter any red passages. 
Notice that even though the optimal plan also avoids the long corridor towards D and collects C first, it also crosses the short passage near a fire (near C) to collect A, which could cause confusion. 
%Essentially, the robot chooses to pass closer to the short red passage to avoid a long detour which incurs a higher navigation cost. 
A key observation here is that the ActiveEXP plan incurs slightly more cost than the optimal plan due to the detour while still being explicable due to the dynamic nature of the human's understanding. 
Without modeling the dynamic change of the human's understanding, it would be difficult to generate plans like ActiveEXP.

\section{Related Work}

% The underlying assumption in earlier work on explicability using model based methods is that the robot has access to the human's model beforehand.
% In reality, such information is difficult to obtain without explicitly querying a human, for his/her expectations of the robot. Also, this implies that the human's expectations of the robot are known apriori and do not change during the course of plan execution. This is unlikely because the human's belief changes while observing the robot which also causes the expectations to change (which is unaccounted in existing methods). While the existing approaches focus on generating task specific plans and therefore, are designed to work on short term based plan solutions, they will undoubtedly fail in long-term interactions due to the dynamically changing expectations of the human. Therefore, maintaining a dynamic human belief in such cases becomes critical for teaming effectiveness. 

Explicable plan generation falls under the umbrella of explainable planning \cite{Chakraborti2017AICI}, which has attracted a significant amount of attention. 
Methods have been proposed under various similar but different notations that include explicability \cite{Zhang2017PlanEA, zakershahrak2018interactive, Kulkarni2016ExplicableRP}, interpretability~\cite{Chakraborti2019ExplicabilityLP, Sreedharan2020ABA}, predictability~\cite{Chakraborti2019ExplicabilityLP}, legibility \cite{Dragan2013GeneratingLM, Kulkarni2019AUF}, and transparency \cite{MacNally2018ActionSF}.  
One of the common features of such methods is the focus on assisting the understanding of different aspects of the plan,
whether it is about the goal, actions, or rationale of the plan. 
% which focus on generating plans that assist in understanding the plans or goals of the executing agent. 

Our work on active explicable planning is closely connected to plan explicability \cite{Zhang2017PlanEA, Kulkarni2016ExplicableRP}, where the aim is to generate behaviors that are expected by the human teammate, assuming that the goal is known. 
This is critical in human-robot teaming tasks since any inconsistency between the robot's behavior and human's expectation can lead to reduced situation awareness and the loss of trust.
In \cite{Kulkarni2016ExplicableRP}, the authors assume that the human's model of expectation is known. It then learns a regression model to predict a distance metric between the robot's plan and the expected plan, which is used to choose a plan that is closer to the expected plan. 
Authors in \cite{Zhang2017PlanEA} relax the assumption of a known human model. 
They learn the human's expectations via a sequential labeling scheme. 
A more principled way to relax this assumption is to maintain a belief of the model of expectation, similar to \cite{Sreedharan2020ABA}.
However, these prior works all assume that the human's model of expectation remains fixed throughout the plan.  

The dynamic modeling of the human's understanding of the robot in our work has close connections to intent, plan and goal recognition \cite{Ramrez2010ProbabilisticPR, Kautz1986GeneralizedPR, Charniak1993ABM, Levine2014ConcurrentPR}, where observations inform the recognition. It more generally involves the dynamic modeling of mental state,  such as trust \cite{ElSalamouny2009HMMBasedTM, Xia2019TrustMI, Liu2012ModelingCA}, except that the mental state to be modeled here is the human's understanding of the robot's domain model (i.e., the model of expectation).
While it may be tempting to apply a POMDP framework~\cite{kaelbling1998planning}, 
it is impractical as the observation function and the explicability score (to be discussed soon) in our problem are plan-context dependent, and hence the POMDP will be computationally expensive to solve.

% Even though POMDP~\cite{Chen2016POMDPliteFR}, \cite{Pineau01ahierarchical}, \cite{7139219} can be used to consider decision making, 
% we cannot directly apply it since the observation function is not just a function of the state. 
% Also, POMDP solutions are computationally expensive.

\section{Problem Formulation}
In this work, we represent domain models using 
%the definition of Classical Planning \cite{russel2010} represented by the 
the PDDL language \cite{DBLP:journals/corr/abs-1106-4561}. A model $\mathcal{M}$ = (($\mathcal{F}$, $\mathcal{A}$), $\mathcal{I}$, $\mathcal{G}$) where $\mathcal{F}$ and $\mathcal{A}$ are a finite set of fluents and actions, respectively. Also, $\forall a$ $\in$ $\mathcal{A}$, an action $a$ is specified as a set of preconditions Pre($a$), add effects Add($a$), and delete effect Del($a$),
which are subsets of $\mathcal{F}$. 
 Each action is also associated with a cost c($a$).
 Let $S$ be the set of states where $s$ $\in$ $S$ is a unique instantiation of the set of fluents $\mathcal{F}$. The initial state $\mathcal{I}$ $\in$ $S$ and the goal state $\mathcal{G}$ $\in$ $S$. We assume that the initial state and goal states are known to the human and robot and both the human and robot are rational agents. 
We first define a list of terms used throughout the paper:
\begin{itemize}
        \item $M_R$ is the model for generating the robot's behavior.
        \item $M_H$ is the human's model of expectation (i.e., an understanding of the robot's model $M_R$).
        \item $b_0$ ($M_H$) is the initial belief of the human's understanding of the robot. 
        \item $I$ is the initial state.
        \item $G$ is the goal state.
        \item c($\pi$) is cost of the plan $\pi$
        \item $\gamma$ is a weighting factor.
    \end{itemize}
    
\begin{definition}
    An \textbf{explicable planning} problem \cite{Kulkarni2016ExplicableRP, Chakraborti2019ExplicabilityLP} is defined as:
    given a tuple $\mathcal{P_E}$ = $\langle$ $M_R$, $M_H$, $I$, $G$ $\rangle$, find a plan $\pi_\mathcal{E}$ that satisfies the following:
    \begin{equation}
       \pi_{\mathcal{E}} = \argminpi_{\pi_{M_R}} \hspace{2pt} \phi(\pi_{M_R}, \pi_{M_H)} +  \hspace{2pt} \gamma c(\pi_{M_R}) 
    \label{ExpPlan}
\end{equation}

    where, $\phi$ is the explicability distance that captures the difference between the plans $\pi_{M_R}$ and $\pi_{M_H}$, which are generated by $M_R$ and $M_H$, respectively.
\end{definition}

To extend the formulation above to a dynamic human model,
% The problem with this formulation is that it assumes a static human model $\mathcal{M_H}$. Moreover, such a model is not easily available to the robot. To address this issue 
we define the problem of active explicable planning.

\begin{definition}
    An \textbf{active explicable planning} problem is defined as: given a tuple $\mathcal{P_{E_A}}$ = $\langle$ $M_R$, $b_0$ ($M_H$), $I$, $G$ $\rangle$, find a plan that satisfies the following:
\begin{equation}
    \pi_{\mathcal{E_A}} = \argmaxpi_{\pi_{M_R}} \hspace{2pt} \mathcal{E_A}(\pi_{M_R}) - \gamma \hspace{2pt} c(\pi_{M_R}) 
    \label{ActiveExpPlan} 
\end{equation} 
    where, $\mathcal{E_A}$ is the active explicability score of a plan (which we will explain in the next section).
\end{definition}

Again, the two main differences between an explicable planning problem $\mathcal{P_E}$ and an active explicable planning problem $\mathcal{P_{E_A}}$ are: 1) $\mathcal{P_E}$ assumes a fixed human model  whereas $\mathcal{P_{E_A}}$ considers that this model may change dynamically during a plan execution, and 2)  $\mathcal{P_{E_A}}$ operates on a belief or a distribution of possible models.

To address this problem, we assume that the human is a Bayesian rational (noisily rational)~\cite{baker2011bayesian} thinker. 
This allows us to model how the human's understanding of the robot may change as observations are made about the robot. 
We also assume that the model space $M$ is known
and both $M_R$ and $M_H$ belong to $M$.
Next, we introduce a few notions that will be used later. Given a pair \{$I$, $G$\}, a candidate plan space $\Pi_i$ = $\{{\pi}_1, {\pi}_2,..., {\pi}_p\}$ is a finite set of $p$ plans generated by a possible model $M_i$ $\in$ $M$. All plans in $\Pi_i$ are bounded by a cost threshold $\zeta$ $\cdot$ c(${\pi}^*$), where $\zeta$ $\geq$ 1 and c(${\pi}^*$) is the cost of the optimal plan for $\{I, G\}$ in $M_i$. The factor $\zeta$ can be arbitrarily set. In our evaluation, we set the cost threshold to be slightly above $1$. %$1.3 1.4$.
%greater than the optimal cost at a given time step. 
% \vspace{2pt} \newline

\section{Methodology}

In explicable planning, the objective is to trade off the plan cost with explicability. 
In this work, we extend explicable planning to consider how the change in human's belief influences the expectations, as observations are made about the robot's actions.

\begin{figure}
    \centering
    \includegraphics[scale=0.34]{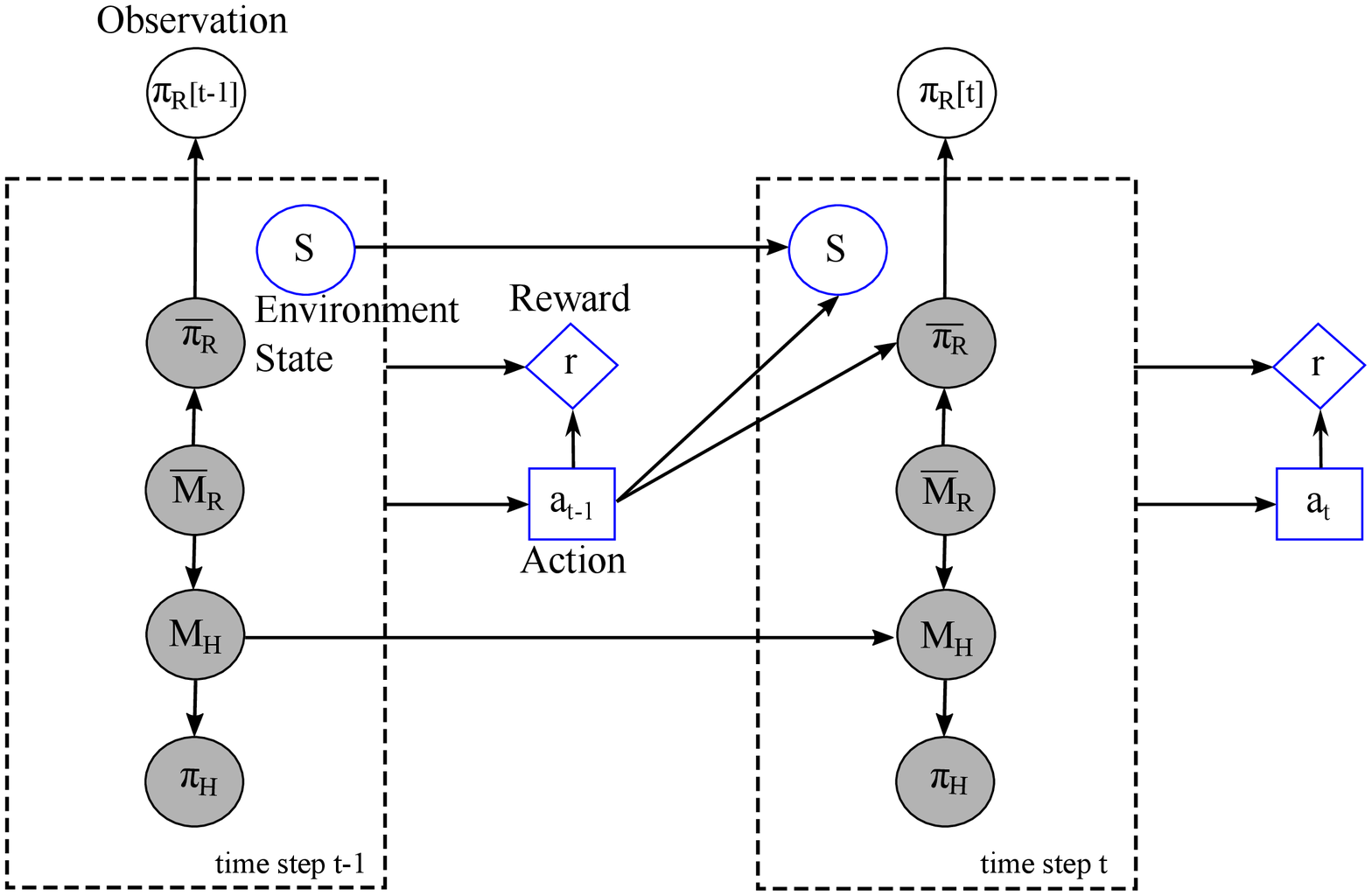}
    \caption{Influence Diagram for active explicable planning for modeling dynamic human belief $M_H$. In this model, we assume that the human uses $M_H$ to generate expectations of the robot's behavior. Furthermore, the human also maintains a temporary robot model $\widebar{M}_R$ (different from $M_R$) to capture the changes to be made to the human's model. These changes are introduced as a result of observing the robot's current action, when it is contrasted to the plan introduced by $\widebar{M}_R$.}
    \label{fig:infdig}
    % \vskip{-10pt}
\end{figure}

\subsection{Active Explicability}
Instead of relying on a distance metric to quantify plan explicability as in 
~\cite{Zhang2017PlanEA, Kulkarni2016ExplicableRP, Chakraborti2019ExplicabilityLP},
we adopt a Bayesian formulation, similar to some plan interpretibility measures in \cite{Sreedharan2020ABA}.
This allows us to define plan explicability measure in a principled way by directly comparing plans. 
However, the limitation is that it will not work when the human's and the robot's model spaces are different. 
We will delay further discussions on this to future work. 
More specifically, the explicability of a plan $\pi_R$ in our work is defined as the probability of the plan $\pi_R$ being in a set of candidate plans that can be generated by the human's model $M_H$. Intuitively, a plan that is more likely to be generated by the human's model is more explicable. The explicability of a plan $\pi_R$ is given by:

\begin{equation}
    \mathcal{E}(\pi_R) = f(\pi_R , M_H) = P (\pi_R \in \Pi_H | M_H)
    \label{exp1} 
\end{equation}

where $\Pi_H$ is the set of candidate plans in $M_H$ with cost less than $\zeta$ $\cdot$ c(${\pi}^*$). 
%i.e.,  $\forall$ $\pi_i \in \Pi_H$, c($\pi^*$) $\leq$ c($\pi_i$) $\leq$ $\zeta$ $\cdot$ c(${\pi}^*$). 
We further assume the probability of such plans being considered follows a noisy distribution (e.g, Boltzmann) based on the cost difference with $\pi^*$.
%However, in the above formulation the human model $M_H$ is assumed to be known. 
When $M_H$ is unknown, we consider a belief distribution over all possible models in $M$, represented by $b(M_H)$. The explicability of a plan $\pi_R$ is defined as follows:

\begin{multline}
    \mathcal{E}(\pi_R) = f(\pi_R , b(M_H)) \\
    = \sum_{M_H} P (\pi_R \in \Pi_H | M_H ) \cdot b(M_H)
    \label{exp2}  
\end{multline}

Furthermore, to capture that $b(M_H)$ can change over time, we define the active explicability of $\pi_R$ to be the average of action explicability measures at each plan step as in \cite{Zhang2017PlanEA}:
 
\begin{multline}
    \mathcal{E_A}(\pi_R) =  \frac{1}{T} \sum_{t} f(\pi_R[t] , b^t(M_H^t)) \\
    = \frac{1}{T} \sum_{t}  \sum_{M_H^t} P (\pi_R[t] \in \Pi_H^t  | M_H^t) \cdot b^t(M_H^t)
    \label{activeExp1} 
\end{multline}

where $\pi_R[t]$ represents the action at step $t$ in $\pi_R$ and $\Pi_H^t$ represents the set of candidate plans in $M_H^t$ 
%such that $\pi_R$[t] is the $t^{th}$ action in the plans 
generated at time $t$ with initial state $S^t$ (i.e., the resulting state from $I$ after executing the first $t-1$ actions in $\pi_R$) and goal state $G$.
%where ($S^t$ and $G$) $\in$ $S$. 
Notice that since new plans are generated at every time step with initial state as $S^t$, the cost of the optimal plan also changes respectively. 
This assumes that the human will change his/her expectation of the robot's plan based on the current state after observing some robot actions. 
In such a case, if the robot's next action matches with the first action of an expected plan from the current state, the action is considered to be consistent with that plan. 
Note that we slightly abuse the notation here and use $\in$ to check this consistency. 
%Essentially, this means that the action at step $t$ must be explicable in the human model. 
To further clarify the notation, we incorporate a delta function $\delta (\pi_R[t] \in \pi_H^t)$ to indicate whether an observation $\pi_R[t]$ matches the first action of the candidate plan in $\Pi_H^t$. Consequently, we rewrite the equation above as:

\begin{equation}
     \mathcal{E_A}(\pi_R) = \frac{1}{T} \sum_{t} \sum_{M_H^t} \sum_{\pi_H^t} \delta (\pi_R[t] \in \pi_H^t) P( \pi_H^t  | M_H^t ) \cdot b^t(M_H^t)
    \label{activeExp2}
\end{equation}

where, $P( \pi_H^t  | M_H^t )$ is the likelihood function described by a Boltzmann distribution based on cost difference given by:

\begin{equation}
    \centering
    P ( \pi_H^t |  M_H^t ) \propto \hspace{2pt} \epsilon^{ - \beta  \times  \{c(\pi_H^t) - c(\pi^*)\} }
    % \label{Pi_H}
    \label{Pi_likelihood}
\end{equation}

In theory, a plan is the most explicable when it maximizes the explicability score without considering the cost. However, in practice, it is desirable to choose a plan that balances the explicability score with the cost of a plan \cite{Zhang2017PlanEA}. Similarly, an active explicable plan $\pi_{\mathcal{E_A}}$ is defined as a plan that maximizes a weighted sum of active explicability score and plan cost as described in Eq. \ref{ActiveExpPlan}.

\subsection{Human Belief Update}
The key to computing  Eq. \ref{activeExp2} and in turn Eq. \ref{ActiveExpPlan} is $b^t(M_H^t)$.
To simplify the problem, we assume that the environment state and robot action are observable and that the robot actions are deterministic. 
$\pi_R$ = $\pi_R[1],..\pi_R[t]..$ is the sequence of actions that are observed by the human at the respective time step. 

Consider the dynamic belief network in Fig. \ref{fig:infdig}. There are two factors that influence the human's belief when an observation is made at time $t$: the prior belief of the human model $M_H^{t-1}$ and the belief of a temporary model of the robot $\widebar{M}_R^t$ (which is also used to derive the robot's behavior) upon observing an action $\pi_R[t]$. 
The idea behind considering this temporary model variable is to capture the models that are consistent with the observation made,
which inform updates to the human's belief.
For example, consider the human expects the robot to be only capable of walking but at some point observes the robot fly. In such a case, $\widebar{M}_R^t$ would be consistent with all the candidate models where the robot can fly.

The belief of the human model at a given time $t$, $b^t(M_H^t)$ can be computed using the Forward algorithm derived below:
\begin{equation}
    b^t(M_H^t) = P(M_H^t | \pi_R[1:t] )
    \label{fwd1}
\end{equation}

Given the graphical model in Fig. \ref{fig:infdig}:
\begin{multline}
    b^t(M_H^t) = \sum_{M_H^{t-1}} \sum_{\widebar{M}_R^t} \sum_{\widebar{\pi}_R^t} \sum_{\pi_H^t} P(M_H^t, M_H^{t-1}, \widebar{M}_R^t, \\
    \widebar{\pi}_R^t, \pi_H^t | \pi_R[t], \pi_R[1:t-1] )
    \label{fwd2}
\end{multline}

Expand the above equation using the chain rule and simplify using conditional independencies we get:

\begin{multline}
    b^t(M_H^t) = \sum_{M_H^{t-1}} P(M_H^{t-1} | \pi_R[1:t-1]) \sum_{\widebar{M}_R^t} P(\widebar{M}_R^t) \\
    P(M_H^t | M_H^{t-1}, \widebar{M}_R^t) \sum_{\widebar{\pi}_R^t} P(\widebar{\pi}_R^t | \widebar{M}_R^t) \delta (\pi_R[t] \in \widebar{\pi}_R^t) \\
    \sum_{\pi_H^t} P(\pi_H^t | M_H^t)
    \label{fwd3}
\end{multline}

where, $P(M_H^{t-1}$ $|$ $\pi_R[1:t-1])$  is nothing but $b^{t-1}(M_H^{t-1})$. 
The term $\sum_{\pi_H^t}$ $P(\pi_H^t | M_H^t)$ sums to $1$ so it does not influence the belief update.
Thus,
\begin{multline}
    b^t(M_H^t) = \sum_{M_H^{t-1}} b^{t-1}(M_H^{t-1}) \sum_{\widebar{M}_R^t} P(\widebar{M}_R^t) \\
    P(M_H^t | M_H^{t-1}, \widebar{M}_R^t)
    \sum_{\widebar{\pi}_R^t} P(\widebar{\pi}_R^t | \widebar{M}_R^t) \delta (\pi_R[t] \in \widebar{\pi}_R^t)
    \label{fwd4}
\end{multline}

where the likelihood function of $\widebar{\pi}_R^t$ is similar to that of $\pi_H^t$ described previously.

\section{Evaluation}
In order to evaluate our approach, we run experiments on the IPC Blocksworld domain. The objective here is to demonstrate how dynamically tracking the model can benefit the plan efficiency by reducing the trade-off for explicability. 
To show that this approach is actually effective, we further validate it with the help of human subjects in a Taxi domain. The objective is to test whether the plan generated by our approach is indeed viewed as explicable and preferred, compared to the baseline methods.

For a given domain, we generate the model space $M$ by identifying features that specify the true model $M_R$. These features represent different aspects of the model that the human teammate could have a misunderstanding about (e.g. not know about). If the number of features identified is $k$, then the model space $M$ is a power set of $2^k$ models. 
The plan space $\Pi$ is the union of all plans generated by each of the models bounded by $\zeta$ $\cdot$ c(${\pi}^*$).

\textbf{Dynamic inference:} We create a Dynamic Bayesian Network with nodes and edges as described in the Fig. \ref{fig:infdig}. We initialize $M_H$ as a uniform distribution of all models in $M$. $\widebar{M}_R$ being the human's belief of the temporary robot's model is updated when an observation is made, informing the updates to the human's belief $M_H$ as a result of the influence from the observation. We solve the problem by unrolling the resulting DBN and use Forward inference algorithm to infer the belief updates after sequential observations. We compute the active explicability score of a plan $\pi_R$ using Eq. \ref{activeExp2}.   

For synthetic data, we perform the following evaluations:
(1) When the human's belief $M_H$ is carried forward from one task to another, the belief update process in Eq. \ref{fwd4} will converge to the true robot model $M_R$. 
% This is assuming the robot is rational and the human is noisily rational.
(2) Robustness of the process against noisy observations.
(3) Efficiency of the approach by comparing the Active Explicability score and the cost of the plans generated.
For the human study we aim to validate the following two hypotheses:
\newline
$H1 =$ \textit{"Human's belief can dynamically change throughout a plan."}
\newline
$H2 =$ \textit{"The active explicable plan is comparable to explicable plans while being more cost-efficient."}

\subsection{Evaluation on Synthetic domain}
\begin{figure*}[!htp]
  \centering
  \begin{tabular}{ c @{\hspace{20pt}} c @{\hspace{20pt}} c}
    \includegraphics[width=.55\columnwidth]{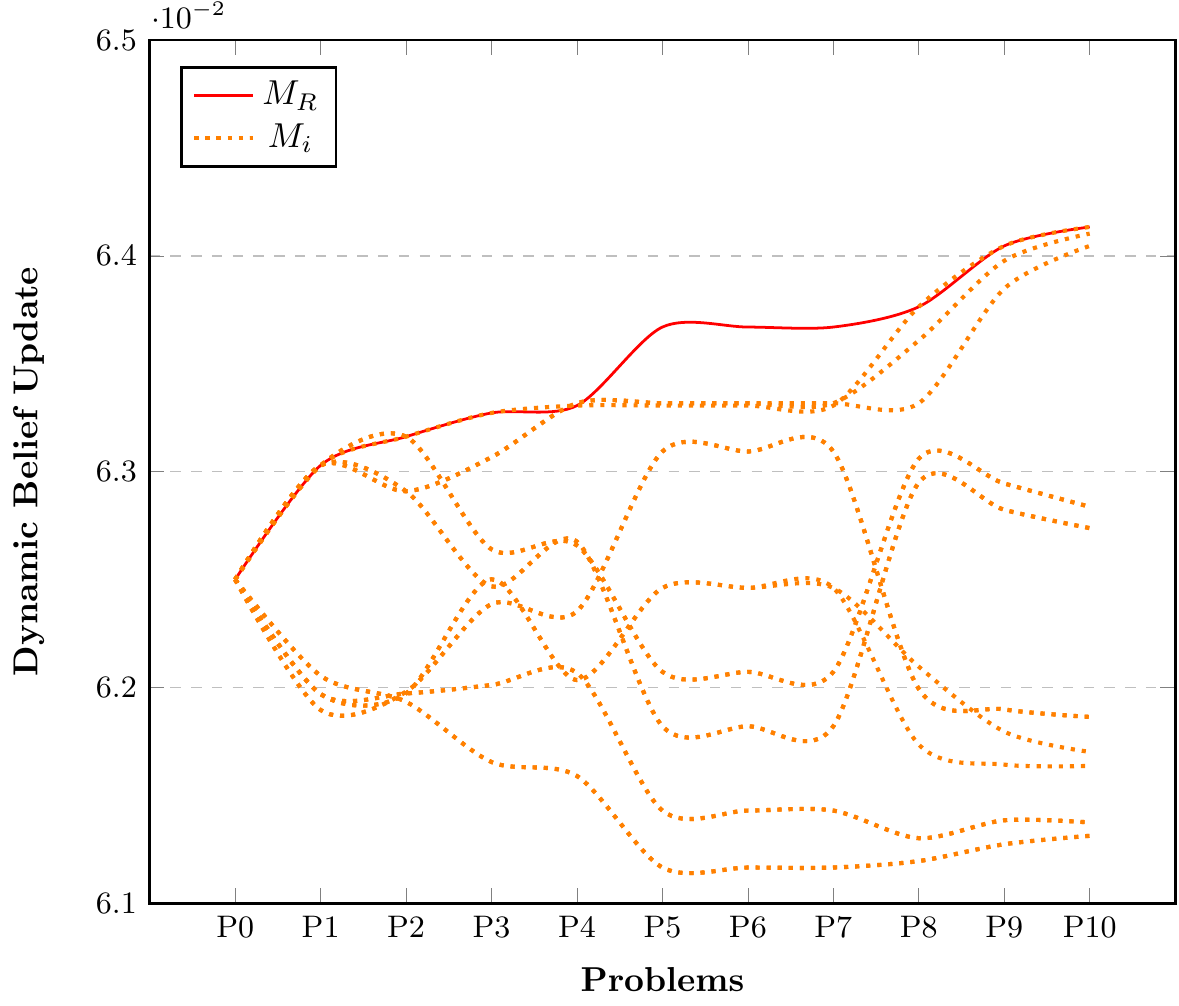}\label{fig:dynamic_belief_update} &
      \includegraphics[width=.55\columnwidth]{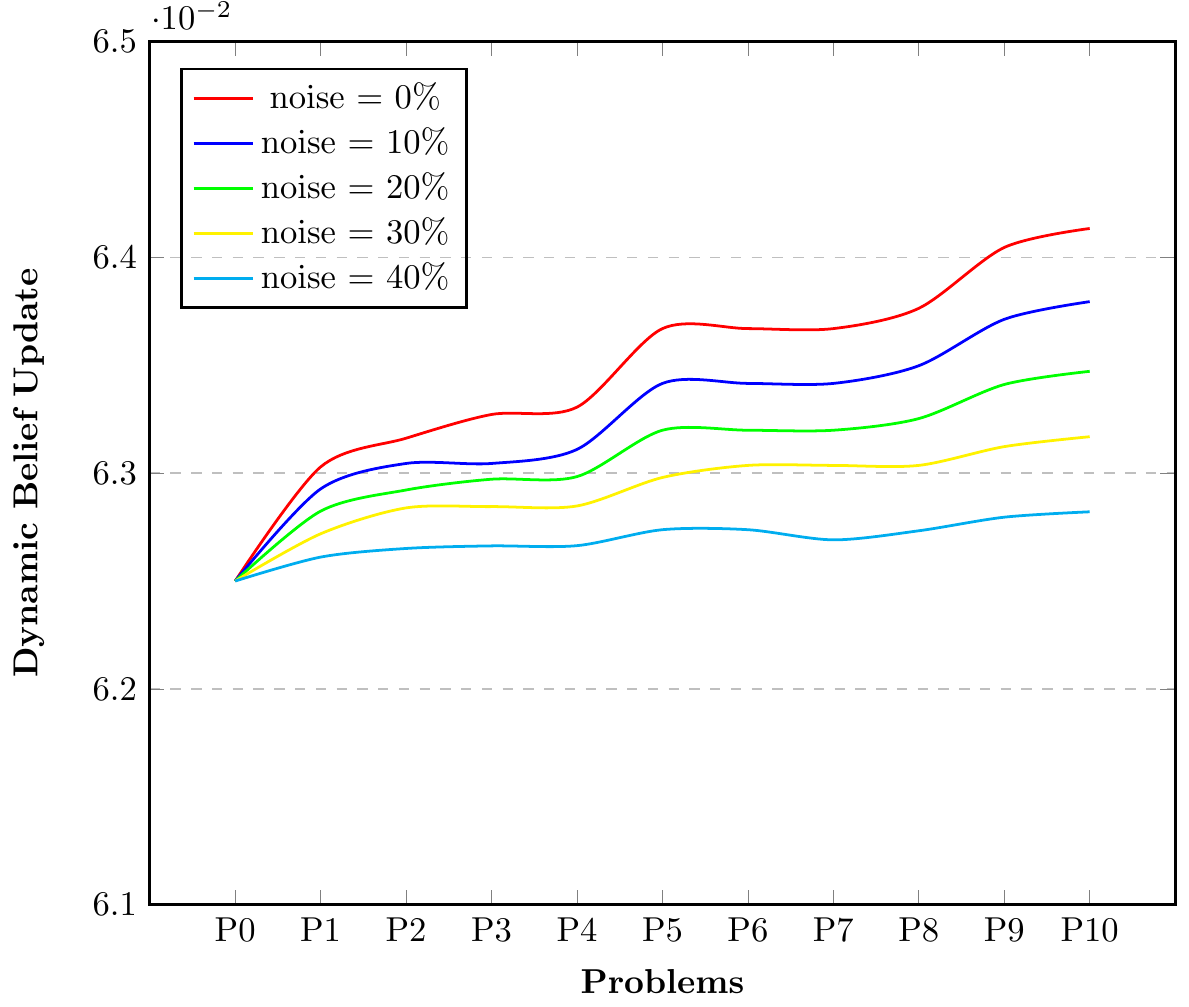}\label{fig:noise_tolerance} &
      \includegraphics[width=.55\columnwidth]{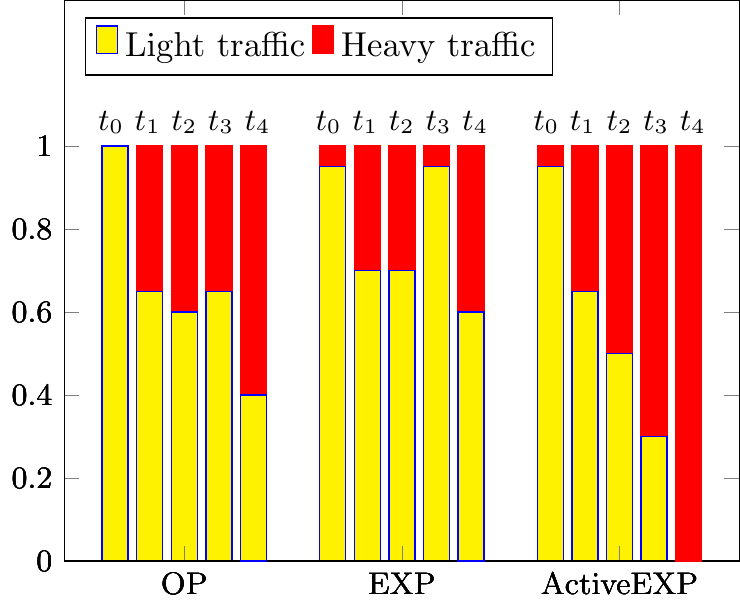}\label{fig:combined_cost} \\
    \small (a) &
      \small (b) & 
        \small (c)
  \end{tabular}
  \medskip
  \vskip-10pt
\caption{  
    (a) Dynamic belief update of all models in $M$ for the Blocksworld. The bold line represents the belief update for the true robot model $M_R$. 
    (b) Belief update of the robot's true model $M_R$ at different noise levels of the observation model.
    (c) Traffic situation estimated by subjects with OP, EXP and ActiveEXP schedules where the ground truth is \textit{heavy traffic}.}
    \label{fig:beliefChange}
\end{figure*}

We use the Blocksworld domain to evaluate how well the dynamic belief update works. We identify $k=4$ features in this domain that could potentially be the factors causing the mismatches between the human's belief and the robot's true model. As a result, $2^4 = 16$ possible models are obtained.
To test (1), we run our approach on 10 different problems with different initial $I$ and goal $G$ states and record the belief update for each of the $16$ models. For the first problem, the human's belief is initialized to be a uniform distribution while for the subsequent problems it is initialized with the updated human's belief from the previous problem.
We do so in support of the intuition that the human carries his/her belief from one problem to another. From the graph in Fig. \ref{fig:beliefChange}. (a), we observe a gradual rise in the belief of the robot's true model $M_R$ starting from a uniform distribution. The belief update is always consistent with the true model albeit being a bit slow. This is 
due to the observed action being generated by at least one plan in every model due to high similarities among the models in $M$.

 Our approach heavily depends on the observations that could affect the dynamic belief update towards the true robot model. In order to investigate this, we measured the belief update for the true robot's model $M_R$ for different noise levels between $0\%$ to $40\%$. \textcolor{black}{We found that the change in belief update for the true robot's model $M_R$ dampens as the noise level increases.}
From Fig. \ref{fig:beliefChange}. (b) we can observe this change. As expected, we found that even with higher noise levels the model updates towards the true robot model.
%however minuscule the update might be. 

We evaluate (3) by testing our approach on four different problems in the Blocksworld domain. Table \ref{tab:bwdomain-cost-score} shows the costs and active explicability scores of the plans respectively. We can observe that the ActiveEXP plans receive higher $\mathcal{E_A}$ scores than OP and EXP plans. In these samples, the ActiveEXP plans overlap with the OP plans in almost all problems due to the limited number of plans generated within the threshold. However, this is not always the case in general.
%and we can observe this in the Taxi domain shown in Table \ref{tab:TaxiPlans}}.

\begin{table}[!h]
\centering
  \begin{tabular}{|l|l|l|l|l|l|l|}
    \hline
     {} & \multicolumn{2}{c}{OP} & \multicolumn{2}{c}{EXP} & \multicolumn{2}{c|}{ActiveEXP} \\
    \hline
   Problem & cost & $\mathcal{E_A}$ & cost & $\mathcal{E_A}$ & cost & $\mathcal{E_A}$ \\
    \hline
    1 & 4 & 0.637 & 6 & 0.509 & 4 & 0.637 \\
    \hline
    2 & 6 & 0.388 & 7 & 0.375 & 6 & 0.437 \\
    \hline
    3 & 8 & 0.562 & 10 & 0.45 & 8 & 0.562 \\
    \hline
    4 & 8 & 0.674 & 9 & 0.6 & 8 & 0.674 \\
    \hline
    % 5 & 5.5 & 5.5 & 5.5 & 5.5 & 5.5 & 5.5 \\
    % \hline
  \end{tabular}
  \caption{Cost and $\mathcal{E_A}$ scores for OP, EXP and ActiveEXP plans generated for different problems in the Blocksworld.}
  \label{tab:bwdomain-cost-score}
\end{table}

% % Synthetic domain: belief update graphs
% \begin{figure*}[!htp]
%   \centering
%   \begin{tabular}{ c @{\hspace{30pt}} c @{\hspace{30pt}} c}
%     \includegraphics[width=.8\columnwidth]{sections/Graphs/dynamicBeliefUpdate.pdf} &
%       \includegraphics[width=.8\columnwidth]{sections/Graphs/noiseTolerance.pdf}\label{fig:noise_tolerance} & \\
%     \small (a)&
%     \small (b)
%   \end{tabular}
%   \medskip
% \caption{ \textbf{(a)} Dynamic belief update of all models in $M$ for the Blocksworld domain. The bold line represents the belief update for the true Robot model $M_R$. 
% \newline \textbf{(b)} Belief update of Robot's true model $M_R$ for different levels of noisy observation model.}
%     \label{fig:dynamic_belief_update}
% \end{figure*}

% % Standalone combined cost graph
% \begin{figure}
%     \centering
%     \includegraphics[scale=0.85]{sections/Graphs/CombinedCost.pdf}
%     \caption{Traffic situation estimated by subjects on observing plans OP, EXP and ActiveEXP where ground truth is \textit{heavy traffic}.}
%     \label{fig:combined_cost}
%     % \vskip{-10pt}
% \end{figure}

% \begin{figure}
%     \centering
%     \includegraphics[scale=0.5]{sections/Graphs/NoiseObsevartion.pdf}
%     \caption{Belief update of Robot's true model $M_R$ for different levels of noisy observation model.}
%     \label{fig:noise_tolerance}
% \end{figure}

\subsection{Evaluation using Human Study}
We design a Taxi domain to evaluate the effectiveness of our approach with human subjects. In the Taxi domain, a private taxi is required to pick up four guests among \{A, B, C, D, E, F, G, H\} from different locations and drop off at the Convention Center (CC). The subjects are provided with a GPS map with locations of the guests and the taxi at the CC. The subjects are informed that picking up guests that are farther (marked in Red) pays more money than those who are closer (marked in Green). However, the traffic situation (unknown to the subjects) can influence the taxi's decision to travel far away. We also inform the subjects that the taxi has access to the current traffic information that may influence its decision. We induce an initial bias of \textit{light traffic} by informing the subjects that the current time is early morning hours. 
% The subjects can infer the traffic situation (Light/Heavy traffic) by observing the plan sequentially.
The traffic situation (Light/Heavy) is the hidden belief which we capture by querying the subjects at every step after revealing the taxi's schedule sequentially. The subject's task is to monitor the taxi agent's schedule. However, the real traffic situation is \textit{heavy traffic}.

\begin{table}[!htp]
    \centering
    \includegraphics[scale=0.23]{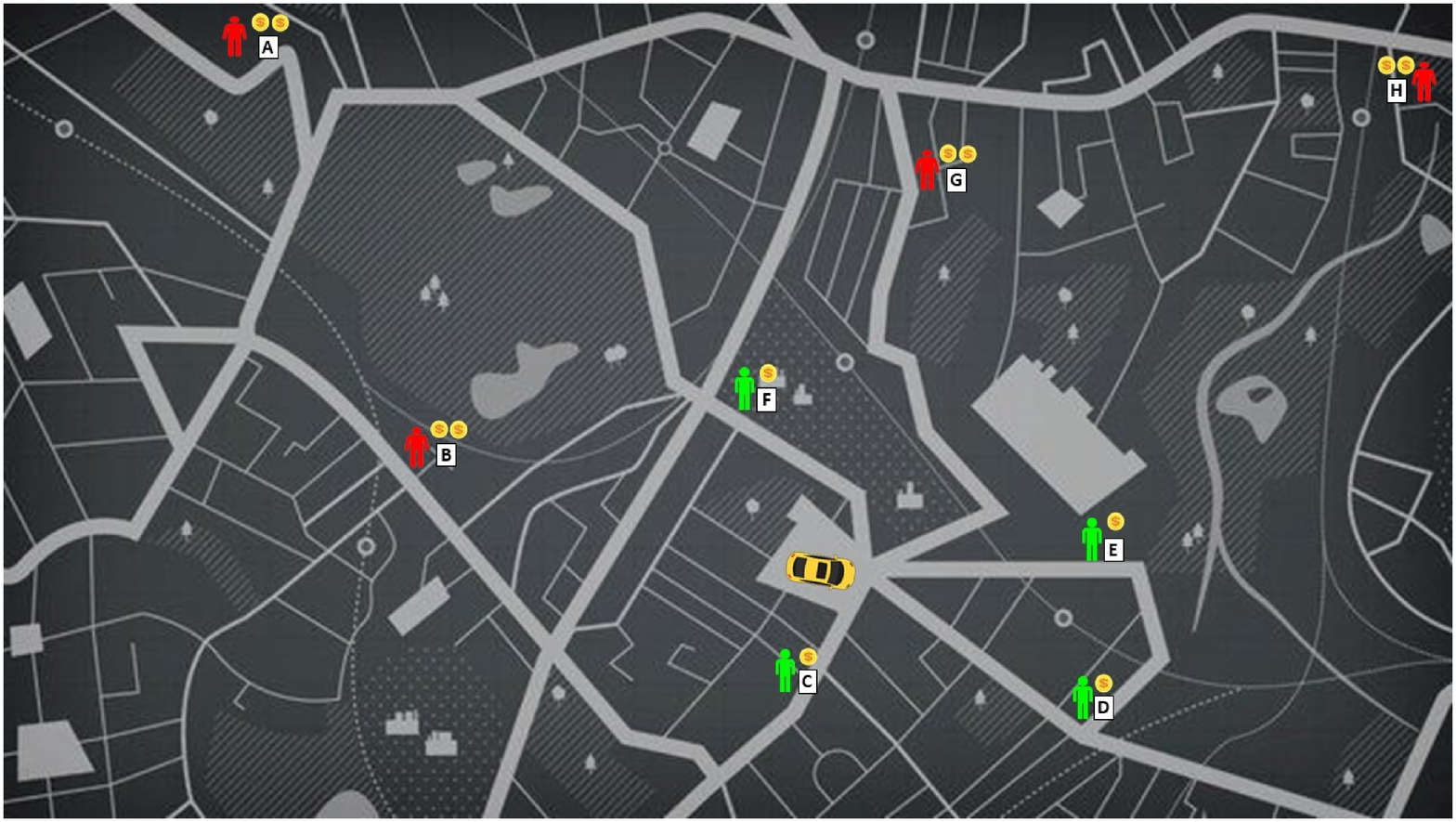} 
    \captionof{figure}{The Taxi domain for human study} 
    \label{fig:layout}
    \vspace{7pt}
    %\resizebox{\linewidth}{!}
    {%
    \begin{tabular}{ |c|c|c|c| }
     \hline
       & OP & EXP & ActiveEXP \\
     \hline
       & Pick \& Drop G & Pick \& Drop G & Pick \& Drop G    \\
 Plan  & Pick \& Drop C & Pick \& Drop B & Pick \& Drop C   \\
       & Pick \& Drop B & Pick \& Drop A & Pick \& Drop D    \\
       & Pick \& Drop F & Pick \& Drop H & Pick \& Drop E     \\
     \hline
   Payoff &  5.2 & -2.4 & 4.2 \\ 
    \hline
   $\mathcal{E_A}$ & 0.123 & 0.161 & 0.143 \\ 
     \hline
    \end{tabular} }
    \captionof{table}{Illustrative Plans OP, EXP and ActiveEXP. Guests marked in red (A, B, G and H) are farther away and pay double than the guests marked in Green (C, D, E and F) who are closer to the Convention Center.}
     \label{tab:TaxiPlans}
     \vspace{-10pt}
\end{table}

In this study, our goal is to compare the ActiveExp plan, the EXP plan and the Optimal plan with the help of human subjects via an Mturk study. We recruited 60 participants (20 for each plan). Before the human subjects made any observations, we asked them about their initial belief of the traffic situation to confirm our induced bias, which is also used as the prior belief. Then, we demonstrated the schedule of the robot sequentially by revealing the next guest to pick up and drop off. We also asked the subjects about their belief of the traffic situation at each step. 

Using this domain, we evaluate three plans ActiveEXP, EXP and OP shown in Table \ref{tab:TaxiPlans} to test $H1$ and $H2$. From Fig. \ref{fig:beliefChange}. (c), we observe that the human belief changes dynamically from light traffic (initial belief) to heavy traffic after observing the taxi's schedule. 

The optimal plan (OP) includes picking up G and B who are relatively closer than other guests marked in red, and C and F marked in green. This plan has the optimal payoff but evaluates to the lowest active explicability score ($\mathcal{E_A}$). Intuitively, this plan does not convey the traffic situation to the observer well since it seems to have exhibited "oscillatory" decisions that point to ambiguous conclusions.
Indeed the OP plan in Fig. \ref{fig:beliefChange}. (c) shows mixed responses from subjects about their belief of the traffic situation. 
The explicable plan (EXP) includes picking up all guests marked in red (G, B, A and H). This plan is the most expected by the observer as it accumulates the highest payoff for the job, given a light traffic condition. Notice that this plan evaluates to the highest $\mathcal{E_A}$ score but is very inefficient given heavy traffic and hence in reality has the lowest payoff. Also, it conveys the wrong idea that there is light traffic as shown in Fig. \ref{fig:beliefChange} which conflicts with the ground truth.
The active explicable plan (ActiveEXP) includes picking up the closest red guest G and three green guests (C, D and E). This plan has a slightly lower $\mathcal{E_A}$ score but a much higher payoff than the EXP plan. More importantly, this plan conveys to the observer that there is heavy traffic as shown in Fig. \ref{fig:beliefChange}. (c) by choosing to pick-up the green guests after picking up G. 
These results show that ActiveEXP is more efficient while still being explicable. Thus, they validate our two hypotheses $H1$ and $H2$.

\begin{figure}
    \centering
    \includegraphics[scale=0.6]{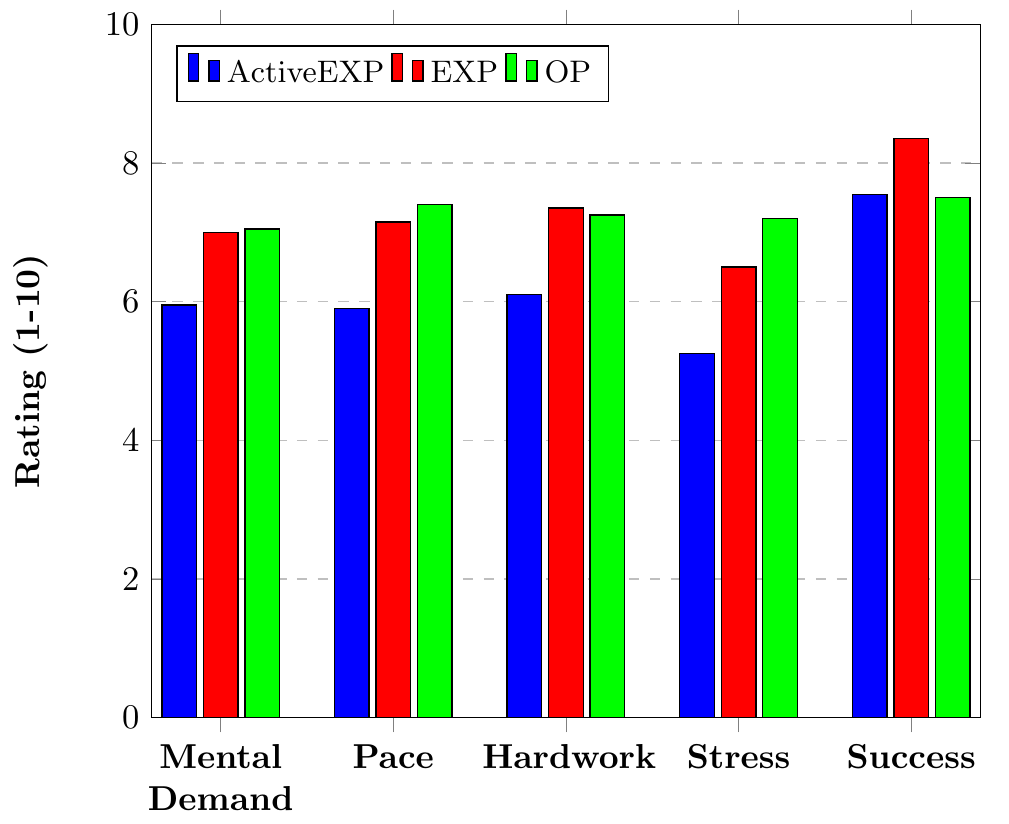}
    \caption{NASA TLX study}
    \label{fig:nasaTLX}
    \vskip-10pt
\end{figure}

In addition, this analysis is confirmed by the subjective results from our study presented in Fig. \ref{fig:nasaTLX}. using NASA TLX. We can see that ActiveEXP performed better than EXP and OP. From statistical analysis (independent t-test), we found that there is a significant difference between ActiveEXP and OP $(p=0.015774)$ and between ActiveEXP and EXP $(p=0.063668)$ at $0.10$ level of significance. 
This result shows that ActiveEXP introduces less cognitive load than the EXP, which is somewhat surprising.
This may be due to the fact that we are constantly expecting changes to occur. We will further analyze it in future work.

% \begin{table}[!htp]
%     \centering
%     {
%     \begin{tabular}{ |c|c|c|c| }
%      \hline
%         Test & t-value & p-value & Result \\
%      \hline
%         $\mu_{AEXP}$ , $\mu_{OP}$ & 1.47462 & 0.075736 & $\mu_{AEXP}$ $=$ $\mu_{OP}$    \\
%         $\mu_{AEXP}$ , $\mu_{EXP}$  & 2.84977 & 0.004059 & $\mu_{AEXP}$ $\neq$ $\mu_{EXP}$    \\
%         $\sigma_{AEXP}$ , $\sigma_{OP}$  & -0.1524 & 0.439981 & $\sigma_{AEXP}$  $=$ $\sigma_{OP}$  \\
%         $\sigma_{AEXP}$ , $\sigma_{EXP}$   & 0.61922 & 0.270389 & $\sigma_{AEXP}$ $=$ $\sigma_{EXP}$    \\ 
%     \hline
%     \end{tabular} }
%     \captionof{table}{Results for Independent t-test for $\alpha = 0.05$}
%      \label{tab:t-test}

% \end{table}

\section{CONCLUSION}

In this paper, we introduced the problem of active explicable planning with dynamic modeling of the human's belief.
It addressed a limitations of the existing work on explicable planning, which assumes a static human belief. We proposed a planning framework based on a Bayesian approach for generating active explicable plans.
We evaluated our method against the existing planning methods and showed that an active explicable plan is more efficient without suffering explicability for human-robot teaming.

\section*{ACKNOWLEDGMENT}
%We thank the anonymous reviewers for their helpful comments which led to significant changes in our work. 
This research is supported in part by the NSF grants 1844524, 2047186, the NASA grant NNX17AD06G, and the AFOSR grant FA9550-18-1-0067.

\addtolength{\textheight}{-12cm}   % This command serves to balance the column lengths

%%%%%%%%%%%%%%%%%%%%%%%%%%%%%%%%%%%%%%%%%%%%%%%%%%%%%%%%%%%%%%%%%%%%%%%%%%%%%%%%

%%%%%%%%%%%%%%%%%%%%%%%%%%%%%%%%%%%%%%%%%%%%%%%%%%%%%%%%%%%%%%%%%%%%%%%%%%%%%%%%



% \section*{ACKNOWLEDGMENT}

% This research is supported in part by the NSF grant IIS-1844524, the NASA grant NNX17AD06G, and the AFOSR grant FA9550-18-1-0067.

%%%%%%%%%%%%%%%%%%%%%%%%%%%%%%%%%%%%%%%%%%%%%%%%%%%%%%%%%%%%%%%%%%%%%%%%%%%%%%%%

\balance
\bibliographystyle{unsrt}
\bibliography{bibtex}

\end{document}